\documentclass{article}

 \usepackage[preprint]{neurips_2025}
\usepackage[utf8]{inputenc} % allow utf-8 input
\usepackage[T1]{fontenc}    % use 8-bit T1 fonts
\usepackage{url}            % simple URL typesetting
\usepackage{booktabs}       % professional-quality tables
\usepackage{amsfonts}       % blackboard math symbols
\usepackage{nicefrac}       % compact symbols for 1/2, etc.
\usepackage{microtype}      % microtypography
\usepackage[table]{xcolor}
\usepackage{hyperref}
\usepackage{amsmath}
\usepackage{subfigure}
\usepackage{graphicx}
\usepackage{subcaption}
\usepackage{multirow}
\usepackage{amssymb}
\usepackage{enumitem}
\usepackage{makecell}
\usepackage{pifont}

\usepackage[most,skins,theorems]{tcolorbox}
\tcbset{
  aibox/.style={
    width=\linewidth,
    top=7pt,
    bottom=2pt,
    colback=blue!6!white,
    colframe=black,
    colbacktitle=black,
    enhanced,
    center,
    attach boxed title to top left={yshift=-0.1in,xshift=0.15in},
    boxed title style={boxrule=0pt,colframe=white,},
  }
}
\newtcolorbox{AIbox}[2][]{aibox,title=#2,#1}

\definecolor{lightgold}{rgb}{1.0, 0.95, 0.8}
\definecolor{light_blue}{RGB}{0, 82, 155}    
\definecolor{light_green}{RGB}{0, 139, 139}  
\definecolor{light_red}{RGB}{204, 0, 0}
\definecolor{rliableblue}{HTML}{77AADD}
\definecolor{table_color}{HTML}{FFB347}

\newcommand{\cmark}{\textcolor{green!60!black}{\ding{51}}}
\newcommand{\xmark}{\textcolor{red}{\ding{55}}}

\newtheorem{template}{Template}
\def\rvo{{\mathbf{o}}}
\def\rvq{{\mathbf{q}}}
\def\gQ{{\mathcal{Q}}}

\title{GUI-G1: Understanding R1-Zero-Like Training for Visual Grounding in GUI Agents}

\author{
Yuqi Zhou\textsuperscript{1}, 
Sunhao Dai\textsuperscript{1}, 
Shuai Wang\textsuperscript{2}, 
Kaiwen Zhou\textsuperscript{2}, 
Qinglin Jia\textsuperscript{2}, 
Jun Xu\textsuperscript{1}  \\
\textsuperscript{1}Gaoling School of Artificial Intelligence, Renmin University of China \\
\textsuperscript{2}Huawei Noah's Ark Lab\\
\texttt{\{yuqizhou,sunhaodai,junxu\}@ruc.edu.cn},\\ 
\texttt{\{wangshuai231,jiaqinglin2,zhoukaiwen2\}@huawei.com}
}

\begin{document}
\maketitle
\begin{abstract}

 Recent Graphical User Interface (GUI) agents replicate the R1-Zero paradigm, coupling online Reinforcement Learning (RL) with explicit chain-of-thought reasoning prior to object grounding and thereby achieving substantial performance gains. In this paper, we first conduct extensive analysis experiments of three key components of that training pipeline: \textcolor{light_red}{input design}, \textcolor{light_green}{output evaluation}, and \textcolor{light_blue}{policy update}—each revealing distinct challenges arising from blindly applying general-purpose RL without adapting to GUI grounding tasks. \textcolor{light_red}{Input design:} Current templates encourage the model to generate chain-of-thought reasoning, but longer chains unexpectedly lead to worse grounding performance. \textcolor{light_green}{Output evaluation:} Reward functions based on hit signals or box area allow models to exploit box size, leading to reward hacking and poor localization quality. \textcolor{light_blue}{Policy update:} Online RL tends to overfit easy examples due to biases in length and sample difficulty, leading to under-optimization on harder cases. To address these issues, we propose three targeted solutions. First, we adopt a \textbf{Fast Thinking Template} that encourages direct answer generation, reducing excessive reasoning during training. Second, we incorporate a box size constraint into the reward function to mitigate reward hacking. Third, we revise the RL objective by adjusting length normalization and adding a difficulty-aware scaling factor, enabling better optimization on hard samples. Our \textbf{GUI-G1-3B}, trained on 17K public samples with Qwen2.5-VL-3B-Instruct, achieves \textbf{90.3\%} accuracy on ScreenSpot and \textbf{37.1\%} on ScreenSpot-Pro. This surpasses all prior models of similar size and even outperforms the larger UI-TARS-7B, establishing a new state-of-the-art in GUI agent grounding. The project repository is available at \url{https://github.com/Yuqi-Zhou/GUI-G1}.

% 一股脑的将GPRO thinking这一套搬到了场景上，缺乏了很细致的消融实验，每一部分没有很好的讨论清楚。作为一篇分析性的工作，直接搬过来看到效果，通过细粒度。缺乏很细致的分析。

% 这一套从这三个最核心的角度去

% 缺一句话从前面抛出来
% rethinking
  
\end{abstract}

\begin{figure}[h]
    \centering
    \includegraphics[width=\textwidth]{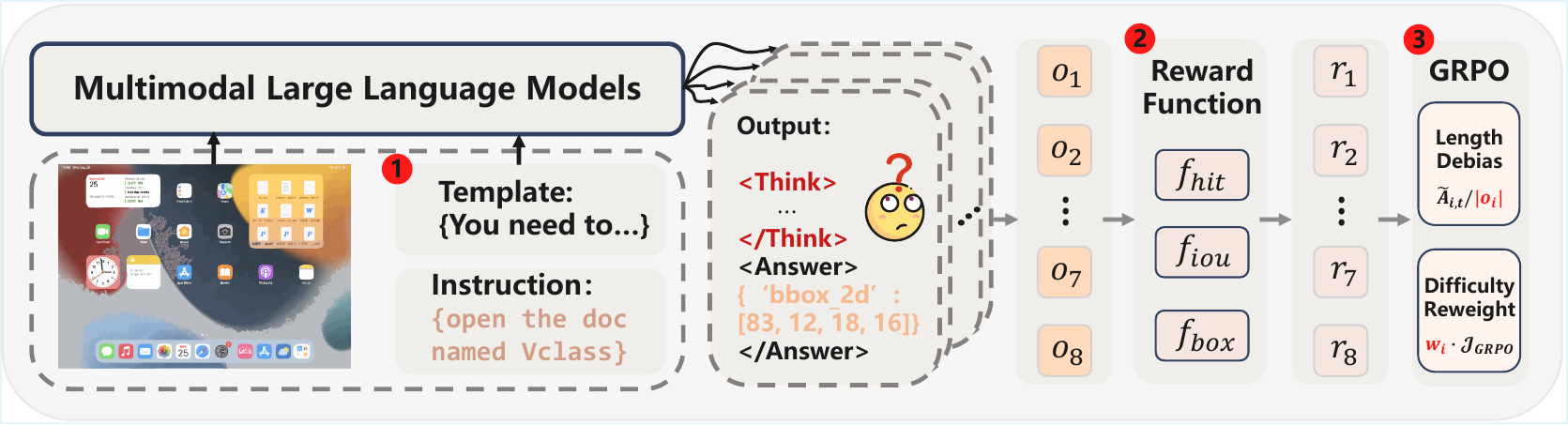}
    \caption{This framework employs the GRPO algorithm for optimization, emphasizing three critical components: \textcolor{light_blue}{input design}, \textcolor{light_green}{output evaluation}, and \textcolor{light_red}{policy update}.}
    \label{fig:intro_fig}
    \vspace{-5pt}
\end{figure}

\section{Introduction}
DeepSeek-R1-Zero~\cite{guo2025deepseek} revolutionizes the post-training pipeline of Large Language Models (LLMs) by introducing the R1-Zero paradigm, which applies RL directly to base LLMs without relying on supervised fine-tuning (SFT) as an intermediate step. Motivated by this approach, recent work in the domain of GUI agents~\cite{liu2025infigui,lu2025ui,xia2025gui} has increasingly adopted RL, particularly the GRPO algorithm~\cite{shao2024deepseekmath}, in order to address two key limitations: (1) SFT requires large-scale, high-quality labeled datasets, resulting in significant computational costs; (2) existing open-source GUI agents trained with SFT often exhibit poor generalization to out-of-domain (OOD) scenarios~\cite{chai2024amex, lu2024omniparser}.

% These R1-based GUI agents have demonstrated strong performance on grounding and action prediction tasks. However, there are many factors that can affect model performance. For instance, existing efforts are all built upon more powerful models such as the Qwen2.5-VL series, while previous state-of-the-art methods, like Os-Atlas~\cite{atlas}, are based on weaker backbones such as Qwen2-VL. This raises an important question: \textbf{how the individual components of the R1-Zero training framework in R1-based GUI agents contribute to model performance?} A more systematic analysis is required to better understand the role of GRPO configurations in this context.

% Furthermore, although prior R1-style agents typically optimize both action prediction and grounding, we argue that grounding is the core capability that underpins effective GUI interaction~\cite{liu2024autoglm}. To facilitate clearer analysis, we focus exclusively on the grounding task~\cite{gou2024navigating,lin2024showui}.

While RL has emerged as a popular choice for training GUI agents in recent work, attributing performance gains solely to the algorithm itself remains nontrivial. These R1-style models often differ in multiple dimensions—including backbone architectures, data sources, and training protocols—making it difficult to isolate the specific contribution of online RL. To better isolate the role of RL, we focus exclusively on the grounding task~\cite{gou2024navigating,lin2024showui}, which we consider the core capability for effective GUI interaction~\cite{liu2024autoglm}. Building on these observations, this work rethinks the role of RL in R1-style GUI agents training by (1) disentangling its algorithmic contributions from other system-level factors, and (2) focusing exclusively on grounding as the reinforcement objective.

% To address the above question, we analyze three key components of the R1-Zero-like training framework: \textcolor{light_red}{the templates}, \textcolor{light_green}{reward functions}, and \textcolor{light_blue}{the GRPO objective}. \textbf{First}, we observe that the grounding performance of the state-of-the-art R1-based model, InfiGUI-R1~\cite{liu2025infigui}, decreases as the thinking increases. This suggests that reasoning may not benefit grounding in GUI agents. \textbf{Second}, we find that commonly used reward functions, such as click-based and IoU-based rewards, lead to opposite forms of reward hacking: the former encourages smaller boxes with higher accuracy, while the latter favors larger boxes to increase IoU. \textbf{Finally}, we identify two biases in the standard GRPO objective: length bias~\cite{liu2025understanding} and difficulty bias. Length bias encourages longer but incorrect responses, which, as previously observed, further degrade grounding performance. Difficulty bias treats all samples equally, hindering the model's ability to learn from more challenging examples. Together, these biases make it harder for the model to learn from difficult samples.

% Each reveals misalignments between model behavior and intended goals.

For this, we decompose the R1-Zero-like training pipeline into three core components: \textcolor{light_red}{input design}, \textcolor{light_green}{output evaluation}, and \textcolor{light_blue}{policy update}. Each reveals distinct challenges arising from blindly applying general-purpose RL without adapting to grounding tasks. \textbf{First}, we observe that the grounding performance of the state-of-the-art R1-style model, InfiGUI-R1~\cite{liu2025infigui}, drops as reasoning increases in Sec.~\ref{sec:analysis_template}, suggesting that reasoning templates may not benefit grounding in GUI agents.
\textbf{Second}, we find that commonly used reward functions based on hit signals or box area lead to opposite forms of reward hacking in Sec.~\ref{sec:analysis_reward}: the former encourages smaller boxes with higher accuracy, while the latter favors larger boxes to increase Intersection over Union (IoU). \textbf{Finally}, we identify two biases in the GRPO objective: length bias~\cite{liu2025understanding} and difficulty bias in Sec.~\ref{sec:analysis_grpo}. Length bias encourages longer but incorrect responses, which, as previously observed, further degrade grounding performance. Difficulty bias treats all samples equally, hindering the model's ability to learn from more challenging examples. Together, these biases make it harder for the model to learn from difficult samples.

% 讲出来length bias在grounding场景里引发了什么额外的危害

To address the above issues, we implement the following improvements. First, we introduce the \textbf{Fast Thinking Template}, which encourages the policy to generate answers directly during training. Second, to counteract the hacking in common reward functions that prefer boxes of different sizes during policy optimization, we propose a box-size-based reward function as a constraint. Finally, we remove the length normalization term from the original GRPO objective as the same in~\cite{liu2025understanding} and introduce a difficulty coefficient for each sample’s loss, allowing the model to receive greater gradients for more challenging samples. The difficulty coefficient is calculated from the relative box size, which serves as a proxy difficulty indicator in the grounding task~\cite{li2025screenspot}.

Building on the above solutions, we train our model, \textbf{GUI-G1-3B}, using Qwen2.5-VL-3B-Instruct and a small (about 17K) set of grounding samples, showing strong performance with limited supervision from public datasets such as UI-BERT~\cite{bai2021uibert} and OS-Atlas~\cite{atlas}. Our model achieves new state-of-the-art performance on GUI grounding benchmarks, with \textbf{90.3\%} accuracy on ScreenSpot~\cite{cheng2024seeclick} and \textbf{37.1\%} on ScreenSpot-Pro~\cite{li2025screenspot}. It surpasses the previous best R1-style GUI agent, InfiGUI-R1~\cite{liu2025infigui}, while requiring significantly less data, fewer output tokens, and fewer training stages. 

In summary, the contributions of this paper are as follows:
\textbf{($1$)} We identify three distinct challenges in the R1-Zero-Like training pipeline of R1-style GUI agents: grounding is harmed by longer reasoning due to grounding's reliance on image tokens; common reward functions induce size-sensitive reward hacking; and GRPO biases agents toward simpler examples due to its objective.
\textbf{($2$)} We further analyze and propose three solutions: a \textbf{Fast Thinking Template} for policy training, a box size–based reward to regularize box size, and a modified GRPO with difficulty weighting and no length normalization.
\textbf{($3$)} Trained on only 17K fully open-source grounding samples, our \textbf{GUI-G1-3B} achieves state-of-the-art performance while using fewer tokens when testing.

\section{R1-Zero-Like Training Paradigm for GUI Grounding}
We begin by explaining how to train Multimodal Large Language Models (MLLMs) in grounding tasks. Given a screenshot $s$ and a textual description $d$, the MLLM is trained to predict the target location $B$, typically represented as a bounding box or a point. Following prior work~\cite{cheng2024seeclick}, we formulate grounding as a language generation task, where the MLLM produces a response $o$ that includes the predicted location as well as other elements such as the reasoning process or objective descriptions in Figure~\ref{fig:intro_fig}. In our implementation, the predicted location is expressed as a bounding box $B_{\text{pred}} = (\hat{x}_1, \hat{y}_1, \hat{x}_2, \hat{y}_2)$, where $x$ and $y$ denote the horizontal and vertical coordinates, respectively. This prediction is evaluated against the ground-truth box $B_{\text{gt}} = (x_1, y_1, x_2, y_2)$.

When RL is applied via the algorithm like GRPO~\cite{shao2024deepseekmath}, a \textcolor{light_blue}{template} is first used to guide the response format, and the model generates $N$ candidate responses $O = \{o_1, o_2, \ldots, o_N\}$. Each response is then evaluated using \textcolor{light_green}{a set of rule-based reward functions}, yielding a reward set $\{r_1, r_2, \ldots, r_N\}$. The relative advantage $A_i$ of each response is computed as:
\begin{equation}~\label{eq:relative_adv}
A_i = \frac{r_i - \texttt{mean}({r_1, r_2, \ldots, r_N})}{\text{std}({r_1, r_2, \ldots, r_N})},
\end{equation}
where $\texttt{mean}$ and $\texttt{std}$ denote the mean and standard deviation of the rewards, respectively. Finally, the policy model is optimized using the \textcolor{light_red}{GRPO objective} with KL-divergence regularization.

\section{How R1-Zero-Like Training Affects Grounding for GUI Agents?}
We first aim to understand R1-Zero-like training paradigm for grounding task in GUI agents by examining three essential components: the \textcolor{light_blue}{input design (template)} (Sec.~\ref{sec:analysis_template}), the \textcolor{light_green}{output evaluation (reward function)} (Sec.~\ref{sec:analysis_reward}), and \textcolor{light_red}{policy update (RL objective)} (Sec.~\ref{sec:analysis_grpo}). Finally, we present our model, \textbf{GUI-G1}, in Sec.~\ref{sec:r1_ground}, where we also summarize and compare our approach with existing R1-style agents to demonstrate its advantages in grounding tasks.

\subsection{Analysis on Template}\label{sec:analysis_template}
Recent R1-style GUI agents have increasingly incorporated explicit reasoning by prompting the model to ``think before action'' \cite{liu2025infigui, lu2025ui,xia2025gui}, as illustrated in Figure~\ref{fig:intro_fig}. For example, InfiGUI-R1\cite {liu2025infigui} uses a \textbf{Slow Thinking Template}. \textbf{While such reasoning-augmented agents achieve strong performance, it remains unclear whether the gains truly arise from the reasoning process itself.} In fact, we find that reasoning is often unnecessary for the grounding task in GUI agents. Before diving into the analysis, we formalize the model’s input and output for consistency across experiments. The input includes an image $s$ and an instruction prompt $\mathbf{t}_{\text{ins}}$, while the output $o$ comprises the reasoning process $\mathbf{t}_{\text{think}}$ and the final answer $\mathbf{t}_{\text{ans}}$, which contains the predicted location $B_{\text{pred}}$. We define the number of reasoning tokens $n_{\text{think}}$ as \textbf{output tokens}, and the tokens derived from the image $n_{\text{img}}$ as \textbf{image tokens}.
The \textcolor{red}{text ratio} is given by $\frac{n_{\text{ins}} + n_{\text{think}}}{n_{\text{img}} + n_{\text{ins}} + n_{\text{think}}},$ where $n_{\text{ins}}$ is the instruction tokens number.

\begin{figure}[h]
    \centering
    \includegraphics[width=\textwidth]{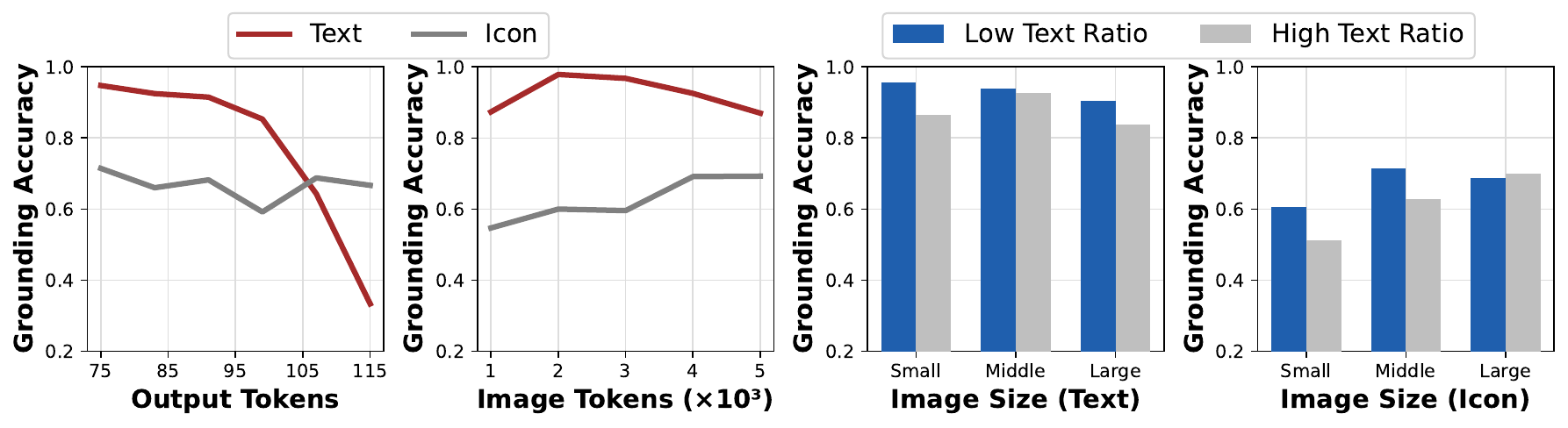}
    \caption{\textbf{(Left)} shows the grounding accuracy under varying numbers of output tokens and image tokens. ``Text'' refers to cases where the target is a textual element, while ``Icon'' refers to image targets. \textbf{(Right)} presents the grounding accuracy on the Text and Icon subsets across different image sizes. Within each group, samples are evenly divided based on their text ratio.}
    \label{fig:analysis_template}
\end{figure}

\textbf{Longer thinking leads to worse grounding performance.} 
While prior work, such as OpenAI-o1~\cite{openai_learning_to_reason} and DeepSeek-R1~\cite{guo2025deepseek} demonstrates that longer reasoning chains can enhance performance on System-2 tasks like mathematics and programming, more recent studies~\cite{li2025think, zhang2025does} have found that introducing intermediate reasoning steps may impair performance in tasks such as image classification and GUI grounding. Building on these observations, we further find that longer reasoning chains consistently degrade grounding accuracy on the ScreenSpot dataset~\cite{cheng2024seeclick}, as shown in Figure~\ref{fig:analysis_template}(Left). This suggests that longer chains are not only unnecessary but can be increasingly detrimental in GUI grounding, especially when the target item to be grounded is text.

\textbf{Grounding benefits from appropriately scaled image tokens rather than from scaled text thinking.}
In Figure~\ref{fig:analysis_template} (Left Middle), we observe that the grounding performance of InfiGUI-R1-3B improves as the number of image tokens increases. This observation raises a central question: \textbf{Is grounding more reliant on image tokens or text tokens?}
To investigate this, we first partition the test samples based on the number of image tokens, ensuring each subset has a comparable level of visual input. Within each subset, we further divide samples into two categories according to their \textcolor{red}{text ratio} and evaluate grounding accuracy for each.
As shown in Figure~\ref{fig:analysis_template} (Right), a higher text ratio consistently correlates with lower grounding performance, indicating that enriching visual content is more effective than injecting additional textual reasoning.

The phenomenon in Figure~\ref{fig:analysis_template} echoes the insight from \textbf{Thinking, Fast and Slow}~\cite{kahneman2011thinking}, which suggests that recognizing familiar visual patterns—such as grounding visual elements—relies on fast, intuitive processes rather than slow, effortful reasoning. Inspired by this, we adopt a \textcolor{light_blue}{Fast Thinking Template} that encourages the model to generate responses without explicit deliberation:

\begin{tcolorbox}[colback=rliableblue!10!white,colframe=black,boxrule=1pt,boxsep=2pt,top=3pt,bottom=3pt,left=2pt,right=2pt]

\begin{template}[\textbf{\emph{Fast Thinking Template}}]
You are a helpful assistant. \textbackslash nUser: Grounding instruction is: \textcolor{red}{\{question\}} Please help to locate and output the bbox coordinates in JSON format.\textbackslash nAssistant:
\end{template}

\begin{template}[\textbf{\emph{Slow Thinking Template}}]
You FIRST think about the reasoning process as an internal monologue and then provide the final answer.\textbackslash nThe reasoning process MUST BE enclosed within <think> </think> tags. \textbackslash n User: The screen's resolution is \textcolor{red}{\{width\}x\{height\}}. \textbackslash nPoint to the UI element most relevant to ``\textcolor{red}{\{question\}}'', output its coordinates using JSON format:\textbackslash n ```json\textbackslash n[\textbackslash n\{\{``point\_2d'': [x, y], ``label'': ``object name/description''\}\}\textbackslash n]```\textbackslash nAssistant:
\end{template}
\end{tcolorbox}

\subsection{Analysis on Reward Function}\label{sec:analysis_reward}
The rule-based reward function introduced in DeepSeek-R1~\cite{guo2025deepseek} exemplifies a simple yet effective approach based on exact match. In grounding tasks, current reward functions for R1-style GUI agents are mainly categorized into \textbf{Hit-based rewards}~\cite{liu2025infigui,lu2025ui,xia2025gui} and \textbf{IoU-based rewards}~\cite{liu2025infigui} in Table~\ref{tab:reward_compare}. Here, $(x_p, y_p)$ is the center of the predicted box, computed as $x_p = (\hat{x}_1 + \hat{x}_2)/2$, $y_p = (\hat{y}_1 + \hat{y}_2)/2$. The Hit-based reward checks whether predicted box center hits within $B_{\text{gt}}$, while the IoU-based reward measures the IoU between $B_{\text{pred}}$ and $B_{\text{gt}}$. While prior work has employed $R_{\text{Hit}}$ and $R_{\text{IoU}}$ as reward signals for grounding-based RL, it remains unclear how these objectives jointly influence training dynamics. To answer this, we implement both types of reward functions for a comparative analysis. The detailed experimental settings and evaluation metrics can be found in Appendix~\ref{sec:anlysis_settings}. Unless otherwise specified, all subsequent analyses follow the same setup.
\begin{table}[h]
\centering
\small
\caption{Comparison of rule-based reward functions and their effects on training dynamics. ``–'' indicates failure to optimize (e.g., $R_{\text{Box}}$ alone).}~\label{tab:reward_compare}
\resizebox{\textwidth}{!}{
\begin{tabular}{l|c|c|c|c|c}
\toprule
\textbf{Reward} & \textbf{Formula} &  \textbf{Driven By} & \textbf{Box Size($\uparrow\downarrow$)} & \textbf{Accuracy($\uparrow\downarrow$)} & \textbf{IoU($\uparrow\downarrow$)} \\
\midrule\midrule
$R_{\text{Hit}}$ & $\mathbf{1}((x_p, y_p) \in B_{\text{gt}})$ & Point Accuracy & $\downarrow$ & $\uparrow$ &  
$\downarrow$ \\
\midrule 
$R_{\text{IoU}}$  & $\text{IoU}(B_{\text{pred}}, B_{\text{gt}})$  & IoU & $\uparrow$  &  $\downarrow$ & $\uparrow$  \\
\midrule
$R_{\text{Box}}$ & $R_{\text{Box}} = \frac{4}{x_{p_1} + x_{p_2} + y_{p_1} + y_{p_2}}$ & Box Size & - & - & - \\
\bottomrule 
\end{tabular}
}
\end{table}

% \textbf{Experimental settings.} We fine-tune all parameters of the Qwen2.5-VL-3B-Instruct model using 900 samples evenly drawn from three domains: mobile (UIBERT~\cite{bai2021uibert}), web, and desktop (OS-Atlas~\cite{atlas}). Training follows the default setup of the VLM-R1 repository~\cite{shen2025vlm}, with 8 rollouts per example and no KL-divergence regularization by default. We report model performance on ScreenSpot (Desktop domain) under different reward functions in terms of grounding accuracy, IoU, and relative box size, as shown in Figure~\ref{fig:analysis_reward}. The relative box size is computed as the sum of the predicted bounding box’s width and height divided by the sum of the corresponding image’s width and height.
\begin{figure}[h]
    \centering
    \includegraphics[width=\textwidth]{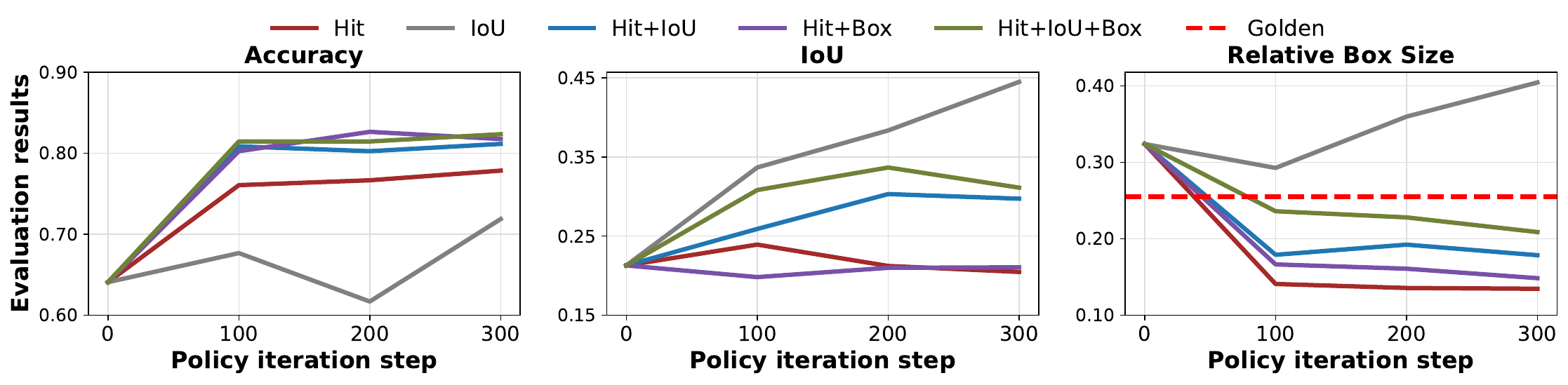}
    \caption{Changes in accuracy (left), IoU (middle), and relative box size (right) across policy iterations during model training on the ScreenSpot dataset.}
    \label{fig:analysis_reward}
\end{figure}

% \textbf{Individually optimizing $R_{\text{Hit}}$ or $R_{\text{IoU}}$ causes opposing types of reward hacking.} As shown in Figure~\ref{fig:analysis_reward} (Left), $R_{\text{Hit}}$ leads to greater improvements in accuracy, with IoU declining in later training stages. Conversely, $R_{\text{IoU}}$ improves IoU but at the cost of reduced accuracy. This reflects the case of \textbf{reward hacking} in GUI grounding tasks, where models adapt to the given objective at the expense of others. Notably, the two metrics capture complementary but competing aspects of model quality: $R_{\text{Hit}}$ evaluates the model's ability to correctly identify the target box, focusing on the precision of localization. However, $R_{\text{IoU}}$ assesses how well the predicted box overlaps with the ground truth, emphasizing the completeness of the localization. These two metrics inherently conflict when optimized independently, highlighting the challenge of designing a balanced reward system.

\textbf{Individually optimizing $R_{\text{Hit}}$ and $R_{\text{IoU}}$ leads to conflicting reward hacking behaviors.} As shown in Figure~\ref{fig:analysis_reward} (Left), optimizing $R_{\text{Hit}}$ improves accuracy but causes IoU to drop in later training. Conversely, optimizing $R_{\text{IoU}}$ enhances overlap quality but reduces accuracy. This illustrates reward hacking in GUI grounding, where models overfit to one objective at the cost of others. These metrics capture complementary yet competing aspects: $R_{\text{Hit}}$ focuses on correctly identifying the target box, while $R_{\text{IoU}}$ measures overlap with ground truth. Their conflict when optimized separately highlights the challenge of designing balanced rewards.

\begin{figure}[h]
    \centering
    \includegraphics[width=0.9\textwidth]{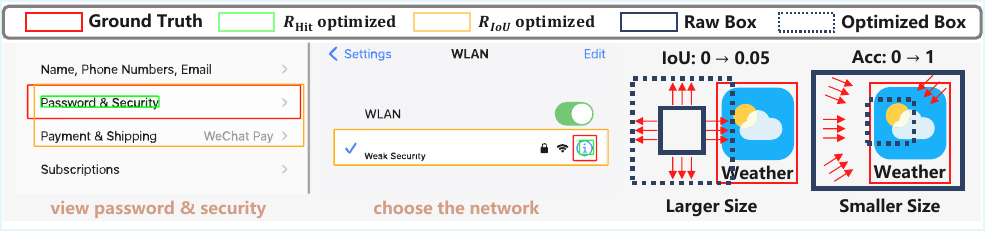}
    \caption{\textbf{(Left)} Two cases with predicted bounding boxes and golden-truth boxes. \textbf{(Right)} Two examples illustrating why $R_{\text{IoU}}$ favors larger boxes, while $R_{\text{Hit}}$ prefers smaller ones.}
    \label{fig:case_study}
\end{figure}

\textbf{GRPO’s sample selection bias toward different box sizes leads to reward hacking.} To investigate the cause of reward hacking, we visualize two cases with predicted bounding boxes in Figure~\ref{fig:case_study} (Left). Models trained with $R_{\text{Hit}}$ tend to produce boxes smaller than the ground truth, while $R_{\text{IoU}}$ leads to significantly larger boxes. This pattern is quantitatively confirmed in Figure~\ref{fig:analysis_reward} (Right), where the relative size of predicted boxes increases over training under $R_{\text{IoU}}$, but decreases under $R_{\text{Hit}}$. Further, as illustrated in Figure~\ref{fig:case_study} (Right), the cause of these opposite trends lies in how GRPO’s sample selection interacts with the reward functions: optimizing $R_{\text{Hit}}$ encourages the model to pick smaller boxes that better capture the core target region, improving accuracy, whereas optimizing $R_{\text{IoU}}$ favors larger boxes that yield higher overlap with ground truth, thus boosting IoU. % This size-driven trade-off creates a seesaw effect between accuracy and IoU.

\textbf{$R_{\text{Box}}$ helps mitigate reward hacking by regularizing box size.}
To address reward hacking, a straightforward solution is to jointly optimize both $R_{\text{Hit}}$ and $R_{\text{IoU}}$. However, as shown in Figure~\ref{fig:analysis_reward}, training may still be dominated by one of the two, resulting in suboptimal balance. To alleviate this, we introduce a new reward function $R_{\text{Box}}$ in Table~\ref{tab:reward_compare}. Here, $x_{p_1} = \frac{1}{1 - |\hat{x}_1 - x_1| / \text{image width}}$, with similar definitions for the other terms. This reward encourages the predicted bounding box to match the ground truth in terms of size. As shown in Figure~\ref{fig:analysis_reward}, incorporating $R_{\text{Box}}$ leads to further improvements in both accuracy and IoU, with predicted box sizes becoming more aligned with the ground truth. We also experiment with using $R_{\text{Box}}$ alone, but the model fails to produce outputs in the correct format. We hypothesize this is because $R_{\text{Box}}$ assigns non-zero rewards even to poorly grounded predictions, encouraging optimization on uninformative samples. Therefore, $R_{\text{Box}}$ should be used in conjunction with $R_{\text{Hit}}$ and $R_{\text{IoU}}$, which directly reflect the evaluation metrics and serve as auxiliary constraints.

\subsection{Analysis on GRPO Objective}\label{sec:analysis_grpo}
Recent approaches to improving GUI agents~\cite{liu2025infigui,lu2025ui,xia2025gui} have adopted RL techniques, such as the GRPO algorithm proposed by DeepSeekMath~\citep{shao2024deepseekmath}. GRPO optimizes the policy $\pi_\theta$ by sampling a set of candidate responses $\{\rvo_i\}_{i=1}^N$ from the old policy $\pi_{\theta_{\text{old}}}$ for each input query $\rvq$, where each response $\rvo_i$ has length $|\rvo_i|$. The policy is updated based on a normalized advantage $\hat{A}_{i,t}$ computed for each token, forming the objective $\mathcal{J}_{\text{GRPO}}(\pi_\theta)$:
\begin{equation}
\footnotesize
\begin{split}
\mathcal{J}_{\text{GRPO}}(\pi_\theta) &= \mathbb{E}_{\rvq \sim p_{\gQ}, \{\rvo_i\}_{i=1}^N \sim \pi_{\theta_{\text{old}}}(\cdot|\rvq)} \\
& \frac{1}{N}\sum_{i=1}^N \textcolor{red}{\frac{1}{|\rvo_i|}} \sum_{t=1}^{|\rvo_i|} \left\{ \min \left[ \frac{\pi_\theta(o_{i,t} | \rvq, \rvo_{i,<t})}{\pi_{\theta_{\text{old}}}(o_{i,t} | \rvq, \rvo_{i,<t})} \hat{A}_{i,t}, \text{clip} \left( \frac{\pi_\theta(o_{i,t} | \rvq, \rvo_{i,<t})}{\pi_{\theta_{\text{old}}}(o_{i,t} | \rvq, \rvo_{i,<t})}, 1 - \epsilon, 1 + \epsilon \right)  \hat{A}_{i,t} \right] \right\},
\end{split}
\label{eq:GRPO-obj}
\end{equation}
where $t$ is the index of the token in the response, $\epsilon$ is a hyperparameter that controls the maximum allowed deviation from the old policy, and $\text{clip}(\cdot, 1 - \epsilon, 1 + \epsilon)$ applies clipping to stabilize training.

\begin{figure}[t]
    \centering
    \includegraphics[width=0.9\textwidth]{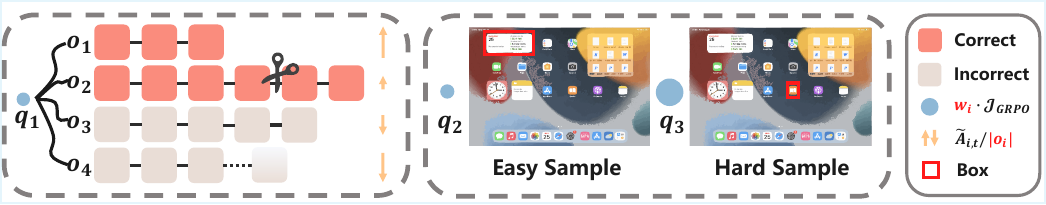}
    \caption{Illustration of the response-level length biases and query-level difficulty biases in GRPO.}
    \label{fig:analysis_grpo}
\end{figure}

\begin{figure}[h]
    \centering
    \includegraphics[width=\textwidth]{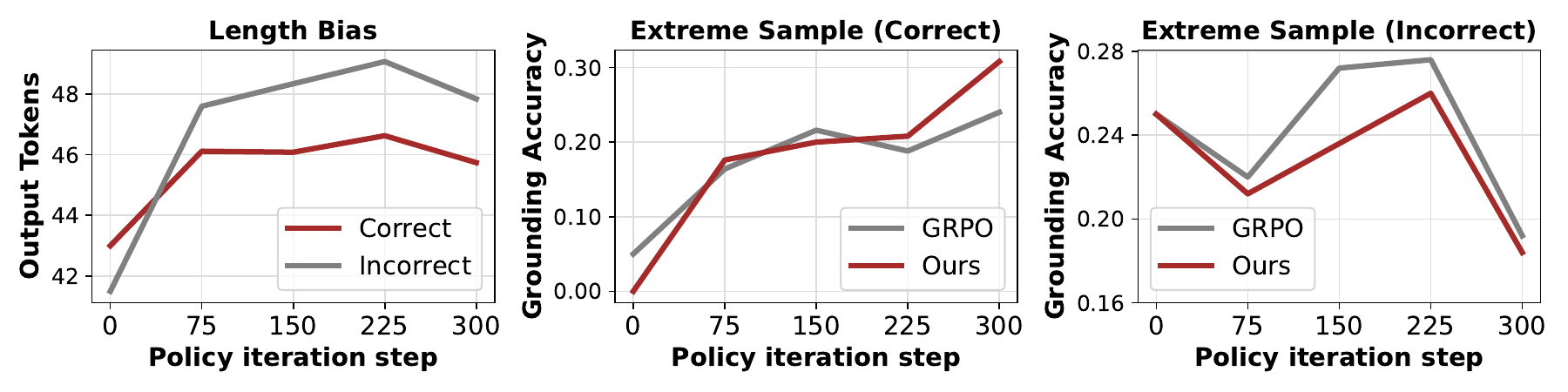}
    \caption{Changes in output length and proportion of extreme samples during policy training.}
    \label{fig:exp_grpo}
\end{figure}

In the setting of GUI grounding tasks, Eq.~\ref{eq:GRPO-obj} introduces two biases (see also in Figure~\ref{fig:analysis_grpo}):
\begin{itemize}[leftmargin=20pt,topsep=-3pt,itemsep=3pt]
    \item\textbf{Response-level length bias~\cite{liu2025understanding}}: It has been observed~\cite{liu2025understanding} that GRPO introduces a length bias: longer responses are preferred among incorrect ones, while shorter responses are favored among correct ones. This arises from dividing the objective $\mathcal{J}_{\text{GRPO}}(\pi_\theta)$ by $\textcolor{red}{|\rvo_i|}$, which amplifies the per-token gradient for shorter responses when the advantage is positive ($\hat{A}_{i,t} > 0$), pushing the policy toward simpler correct outputs. Conversely, it encourages unnecessarily long incorrect answers. As shown in Figure~\ref{fig:exp_grpo} (Left), training gradually results in longer incorrect and shorter correct responses. This trend further harms performance, as longer outputs are shown to degrade accuracy in Section~\ref{sec:analysis_template}. Therefore, length bias in grounding tasks is especially problematic: it not only increases token count but also reduces overall quality. 
    \item\textbf{Question-level difficulty bias}: It has been noted in~\cite{liu2025understanding} that dividing the centered outcome rewards by $\textcolor{black}{\text{std}({r_1, r_2, \ldots, r_N})}$ can lead the model to focus disproportionately on either harder or easier samples. However, we argue that assigning higher weights to harder samples during policy updates is desirable. In grounding tasks, the relative box size of the target can serve as a proxy for task difficulty~\cite{li2025screenspot}. Based on this intuition, we modify the original objective to $\textcolor{red}{w_{q}} \cdot \mathcal{J}_{\text{GRPO}}(\pi_\theta)$, where $w_q$ reflects the difficulty of query $\rvq$. The weight $w_q$ is computed based on the relative box size, where a larger relative size indicates an easier grounding instance. Detailed computation is provided in Appendix~\ref{sec:difficulty_weight}. Multiplying the objective by $w_q$ assigns greater gradients to harder samples, thus encouraging the model to focus on more challenging instances. \textbf{In fact, length bias can also be viewed as a form of difficulty bias, as it guides the model toward generating longer incorrect responses, which exacerbates the difficulty of learning from such examples and indirectly shifts the focus toward easier samples.}
\end{itemize}

\textbf{Experimental Results.} We implement both improvements, with results shown in Table~\ref{tab:grpo_analysis}. Mitigating length and difficulty biases consistently enhances model performance. Figure~\ref{fig:analysis_grpo} (middle and right) further tracks the ratio of extreme samples, where all sampled responses are either correct or incorrect, throughout training. In the middle plot, our method initially lags on easy samples due to their lower weights, but gradually outperforms the original GRPO as these examples are eventually learned. In the right plot, our method maintains a lower proportion of extreme cases on hard samples, indicating that difficulty re-weighting facilitates better learning from challenging instances.

\begin{table*}[h]
\centering
\small
\caption{Evaluation results on ScreenSpot after mitigating length and difficulty biases.}~\label{tab:grpo_analysis}
\resizebox{0.95\textwidth}{!}{%
\begin{tabular}{lcccccccccc}
\toprule
\multirow{2}{*}{\textbf{Training Objective}}  & \multicolumn{3}{c}{\textbf{Mobile}} & \multicolumn{3}{c}{\textbf{Desktop}} & \multicolumn{3}{c}{\textbf{Web}} & \multirow{2}{*}{\textbf{Avg.}} \\ 
\cmidrule(lr){2-4} \cmidrule(lr){5-7} \cmidrule(lr){8-10}
& Text & Icon & Avg. & Text & Icon & Avg. & Text & Icon & Avg. & \\
\midrule
Standard GRPO~\cite{shao2024deepseekmath}                    & 96.5   & 82.4   & 97.8  & 87.6   & 60.7   & 76.3   & 85.0  & 68.4  & 77.3 & \cellcolor{table_color!20}82.3 \\
\quad ${|\rvo_i|} \rightarrow \texttt{Max\_Tokens}$~\cite{liu2025understanding}  & 96.5   & 81.9   & 97.8  & 86.6   & 65.7   & 77.8   & 85.4  & 71.9  & 79.1 & \cellcolor{table_color!20}83.2                 \\
\quad $\mathcal{J}_{\text{GRPO}}(\pi_\theta) \rightarrow w_{p} \cdot \mathcal{J}_{\text{GRPO}}(\pi_\theta)$   & 97.2   & 79.6   & 98.2  & 85.5   & 62.8   & 76.0   & 88.0  & 73.8  & 81.4 & \cellcolor{table_color!20}83.3 \\
\bottomrule
\end{tabular}}
\end{table*}

\subsection{GUI-G1: A Tailored RL Visual Grounding Model}\label{sec:r1_ground}
Based on the above analysis, we identify key limitations in existing training paradigms for grounding tasks. We now summarize our proposed improvements and present a comparison with prior methods.

Our method, GUI-G1, addresses the identified issues through:
\begin{itemize}[leftmargin=20pt,topsep=-3pt,itemsep=3pt]
    \item\textbf{Thinking leads to poorer grounding performance in Sec.~\ref{sec:analysis_template}}: We adopt a template without intermediate reasoning to prevent the policy from generating long thinking during training. 
    \item\textbf{$R_{\text{Hit}}$ and $R_{\text{IoU}}$ cause opposing types of reward hacking in Sec.~\ref{sec:analysis_reward} }: We combine $R_{\text{Hit}}$ and $R_{\text{IoU}}$ as the reward signal and introduce an additional $R_{\text{box}}$ term to regularize predicted box sizes, mitigating reward hacking caused by box size mismatch.
    \item\textbf{Original GRPO introduces length and difficulty biases in Sec.~\ref{sec:analysis_grpo}}: We remove these biases by replacing $|o_i|$ with a constant \texttt{Max\_Tokens}~\cite{liu2025understanding} and by weighting the GRPO objective $\mathcal{J}_{\text{GRPO}}(\pi_\theta)$ with a difficulty coefficient $w_p$.
\end{itemize}

To make the distinctions clearer, Table~\ref{tab:compare} provides a structured comparison between GUI-G1 and existing R1-style GUI agents in grounding tasks.

\begin{table}[h]
\centering
\small
\caption{Comparison of R1-style GUI agents for grounding tasks, focusing on RL, template, reward, and support for length control and difficulty awareness. $\alpha$ and $\beta$ are tunable hyperparameters.}~\label{tab:compare}
\resizebox{\textwidth}{!}{
\begin{tabular}{l|c|c|c|c|c}
\toprule
\textbf{Method} & \textbf{RL} &  \textbf{Template Design} & \textbf{Reward Design} & \textbf{Length Control} & \textbf{Difficulty Aware} \\
\midrule\midrule

\textbf{U1-R1}~\cite{lu2025ui} & GRPO & Thinking Template & $R_{\text{Hit}}$ & \xmark & \xmark  \\
\midrule 
\textbf{GUI-R1}~\cite{xia2025gui} & GRPO & Thinking Template & $R_{\text{Hit}}$ & \xmark & \xmark  \\
\midrule
\textbf{InfiGUI-R1}~\cite{liu2025infigui} & RLOO~\cite{ahmadian2024back,kool2019buy} & Thinking Template & $R_{\text{Hit}}$ + $R_{\text{IoU}}$ & \xmark & \xmark  \\
\midrule 
\textbf{GUI-G1} (Ours) & GRPO & No-Thinking Template & $R_{\text{Hit}}$ + $\alpha R_{\text{IoU}}$+$\beta R_{\text{Box}}$ & \cmark & \cmark \\

\bottomrule 
\end{tabular}
}
\end{table}

\section{Experiments}\label{sec:exp}
In this section, we introduce the experimental setup for training and evaluating our proposed \textbf{GUI-G1-3B} agent. We outline the implementation details, describe the training dataset and evaluation benchmarks, and provide a detailed comparison with state-of-the-art methods.

\begin{table*}[!htp]
    \centering
    \small
    \caption{Performance on ScreenSpot across Mobile, Desktop, and Web. \textbf{Bold} highlights the best results, \underline{underlined} the second-best. ``-'' indicates missing values due to unavailable results in the original paper, unreleased model checkpoints, and unreleased inference code.}
    \label{tab:screenspot}
    \begin{tabular*}{\textwidth}{@{\extracolsep{\fill}}l cccccccc}
    \toprule
    \multirow{1}{*}{\textbf{Model}} & \multirow{1}{*}{\textbf{\#Training Samples}} & \multicolumn{6}{c}{\textbf{Accuracy (\%)}} & \multirow{1}{*}{\textbf{Avg.}} \\
    \cmidrule(lr){3-8}
    & & \multicolumn{2}{c}{Mobile} & \multicolumn{2}{c}{Desktop} & \multicolumn{2}{c}{Web} & \\
    \cmidrule(lr){3-4} \cmidrule(lr){5-6} \cmidrule(lr){7-8}
    & & Text & Icon & Text & Icon & Text & Icon & \\
    \midrule
    \rowcolor{gray!15}
    \multicolumn{9}{l}{\textit{Proprietary Models}} \\
    GPT-4o~\cite{gpt4o} &  -  & 30.5 & 23.2 & 20.6 & 19.4 & 11.1 & 7.8 & \cellcolor{table_color!20}18.8 \\
    Claude Computer Use~\cite{anthropic2024b} & - & - & - & - & - & - & - & \cellcolor{table_color!20}83.0 \\
    Gemini 2.0 (Project Mariner)~\cite{googldeepmind2024} & - & - & - & - & - & - & - & \cellcolor{table_color!20}84.0 \\
    \midrule
    \rowcolor{gray!15}
    \multicolumn{9}{l}{\textit{General Open-source Models}} \\
    Qwen2-VL-7B~\cite{wang2024qwen2} & -  & 61.3 & 39.3 & 52.0 & 45.0 & 33.0 & 21.8 & \cellcolor{table_color!20}42.9 \\
    Qwen2.5-VL-3B~\cite{bai2025qwen2} & - & - & - & - & - & - & - & \cellcolor{table_color!20}55.5 \\
    Qwen2.5-VL-7B~\cite{bai2025qwen2} & - & - & - & - & - & - & - & \cellcolor{table_color!20}{84.7} \\
    \midrule
    \rowcolor{gray!15}
    \multicolumn{9}{l}{\textit{GUI-specific Models (SFT)}} \\
    CogAgent-18B~\cite{hong2024cogagent} & 222M & 67.0 & 24.0 & 74.2 & 20.0 & 70.4 & 28.6 & \cellcolor{table_color!20}47.4 \\
    SeeClick-9.6B~\cite{cheng2024seeclick} & 1M & 78.0 & 52.0 & 72.2 & 30.0 & 55.7 & 32.5 & \cellcolor{table_color!20}53.4 \\
    UGround-7B~\cite{gou2024navigating} & 10M  & 82.8 & 60.3 & 82.5 & 63.6 & 80.4 & 70.4 & \cellcolor{table_color!20}73.3 \\
    OS-Atlas-7B~\cite{atlas} & 13M & 93.0 & 72.9 & 91.8 & 62.9 & 90.9 & 74.3 & \cellcolor{table_color!20}82.5 \\
    ShowUI-2B~\cite{lin2024showui} & 256K  & 92.3 & 75.5 & 76.3 & 61.1 & 81.7 & 63.6 & \cellcolor{table_color!20}75.1 \\
    Aguvis-7B~\cite{xu2024aguvis} & 1M & 95.6 & 77.7 & {93.8} & 67.1 & 88.3 & 75.2 & \cellcolor{table_color!20}84.4 \\
    UI-TARS-2B~\cite{qin2025ui} & -  & 93.0 & 75.5 & 90.7 & 68.6 & 84.3 & 74.8 & \cellcolor{table_color!20}82.3 \\
    \midrule
    \rowcolor{gray!15}
    \multicolumn{9}{l}{\textit{GUI-specific Models (RL)}} \\
    UI-R1-3B~\cite{lu2025ui} & 136 & - & - & 90.2 & 59.3 & 85.2 & 73.3 & \cellcolor{table_color!20}- \\
    GUI-R1-3B~\cite{xia2025gui} & 3K & - & - & {93.8} & 64.8 & 89.6 & 72.1 & \cellcolor{table_color!20}- \\
    GUI-R1-7B~\cite{xia2025gui} & 3K & - & - & 91.8 & 73.6 & {91.3} & {75.7} & \cellcolor{table_color!20}- \\
    {InfiGUI-R1-3B}~\cite{liu2025infigui} & 32K & \underline{97.1} & \underline{81.2} & \underline{94.3} & \underline{77.1} & \textbf{91.7} & \underline{77.6} & \cellcolor{table_color!20}\underline{87.5} \\
    \midrule
    \rowcolor{gray!15}
    \multicolumn{9}{l}{\textit{Ours}} \\
    \textbf{GUI-G1-3B} & 17K & \textbf{98.6} & \textbf{85.8} & \textbf{96.4} & \textbf{80.7} & \underline{91.4} & \textbf{82.3} & \cellcolor{table_color!20}\textbf{90.3} \\
    \bottomrule
    \end{tabular*}
\end{table*}

\textbf{Implementation Details.} Our model is built upon the Qwen2.5-VL-3B-Instruct and trained using the VLM-R1 framework~\cite{shen2025vlm}. The reward function follows the form $R_{\text{Hit}} + \alpha R_{\text{IoU}} + \beta R_{\text{Box}}$, where $\alpha$ is set to 0.25 and $\beta$ to 0.125. We conduct training on 4 NVIDIA H800 GPUs over 3 days, with a global batch size of 32 and a learning rate of $1 \times 10^{-6}$. No KL divergence regularization is applied. Only one training epoch is required.

\textbf{Training Dataset and Evaluation Benchmarks.} We construct a 17K-sample grounding dataset spanning three domains: \textbf{Mobile} (from UI-BERT~\cite{bai2021uibert}), \textbf{Web} (from OS-Atlas~\cite{atlas}), and \textbf{Desktop} (from OS-Atlas, covering Windows, Linux, and MacOS). More details of the training dataset are shown in the Appendix~\ref{sec:training_data}. To ensure data quality, each sample is prompted eight times using Qwen2.5-VL-3B-Instruct, and those with consistently correct or incorrect responses are discarded~\cite{chen2025empirical}. For evaluation, we adopt ScreenSpot~\cite{cheng2024seeclick} and \textbf{ScreenSpot-Pro}~\cite{li2025screenspot}. While ScreenSpot assesses grounding performance across diverse platforms, including Mobile, Web, and Desktop, ScreenSpot-Pro emphasizes more challenging desktop scenarios, featuring high-resolution screens. 

% \textbf{Performance on ScreenSpot.} We compare \textbf{GUI-G1-3B} with a range of state-of-the-art open-source and proprietary GUI agents, using results reported in their original papers. Table~\ref{tab:screenspot} summarize the results on the ScreenSpot benchmark across the Mobile, Desktop, and Web platforms. Our model achieves state-of-the-art performance, outperforming various models, including proprietary systems like Gemini 2.0~\cite{googldeepmind2024}, general open-source models like the Qwen2.5 series~\cite{bai2025qwen2}, SFT-based GUI-specific models like OS-Atlas~\cite{atlas} and UGround~\cite{gou2024navigating}, as well as R1-style models like UI-R1~\cite{lu2025ui}, GUI-R1~\cite{xia2025gui}, and InfiGUI-R1~\cite{liu2025infigui}. Notably, even larger models like UI-TARS-7B~\cite{qin2025ui} are surpassed. Additionally, our approach requires only 17K training samples and does not rely on intermediate reasoning steps, showcasing its efficiency. Notably, our model achieves higher efficiency during inference by using significantly fewer tokens compared to other methods. Detailed results are provided in the Appendix~\ref{sec:output_tokens} Table~\ref{tab:output_tokens}.

 \textbf{Performance Comparison on ScreenSpot.} We compare \textbf{GUI-G1-3B} with a range of state-of-the-art open-source and proprietary GUI agents, using results reported in their original papers. Table~\ref{tab:screenspot} summarizes performance on the ScreenSpot benchmark. \textbf{GUI-G1-3B} achieves state-of-the-art results, outperforming proprietary systems like Gemini 2.0~\cite{googldeepmind2024}, general-purpose models such as the Qwen2.5 series~\cite{bai2025qwen2}, GUI-specific SFT models like OS-Atlas~\cite{atlas} and UGround~\cite{gou2024navigating}, as well as R1-style models including UI-R1~\cite{lu2025ui}, GUI-R1~\cite{xia2025gui}, and InfiGUI-R1~\cite{liu2025infigui}. It also surpasses larger models like OS-Atlas-7B~\cite{atlas}. Despite its strong performance, our model is trained on only 17K samples and requires no intermediate reasoning steps. Moreover, it achieves higher inference efficiency by generating significantly fewer tokens than other methods (see Appendix~\ref{sec:output_tokens}, Table~\ref{tab:output_tokens}).

\textbf{Performance Comparison on ScreenSpot-Pro.} As shown in Table~\ref{tab:screenspot_pro}, \textbf{GUI-G1-3B} achieves competitive performance on the challenging ScreenSpot-Pro benchmark, with an overall average score of 37.1\%. It outperforms the larger UI-TARS-7B model (35.7\%) and significantly surpasses the best-performing R1-based model, InfiGUI-R1-3B (35.7\%). Although both \textbf{GUI-G1-3B} and OS-Atlas-7B use the same training dataset, our model performs worse on the OS subset (16.1\% vs. OS-Atlas-7B's 16.8\%), suggesting that its gains mainly result from post-training that activates pretrained knowledge rather than from task-specific data. This demonstrates the robustness and generalization ability of our approach in real-world scenarios.

\begin{table*}[!htp]
    \centering
    \caption{Comparison of agent models on ScreenSpot-Pro across Text, Icon, and Average task metrics. Best results are shown in \textbf{bold}, with second-best results \underline{underlined}.}
    \label{tab:screenspot_pro}
    \scriptsize
    \setlength{\tabcolsep}{1pt}
    \resizebox{\textwidth}{!}{
    \begin{tabular*}{\textwidth}{@{\extracolsep{\fill}}l *{21}{c}}
    \toprule
    \multirow{3}{*}{\textbf{Model}} & \multicolumn{3}{c}{\textbf{CAD}} & \multicolumn{3}{c}{\textbf{Development}} & \multicolumn{3}{c}{\textbf{Creative}} & \multicolumn{3}{c}{\textbf{Scientific}} & \multicolumn{3}{c}{\textbf{Office}} & \multicolumn{3}{c}{\textbf{OS}} & \multicolumn{3}{c}{\textbf{Avg.}} \\
    \cmidrule(lr){2-4} \cmidrule(lr){5-7} \cmidrule(lr){8-10} \cmidrule(lr){11-13} \cmidrule(lr){14-16} \cmidrule(lr){17-19} \cmidrule(lr){20-22}
    & Text & Icon & Avg. & Text & Icon & Avg. & Text & Icon & Avg. & Text & Icon & Avg. & Text & Icon & Avg. & Text & Icon & Avg. & Text & Icon & Avg. \\
    \midrule
    \rowcolor{gray!15}
    \multicolumn{22}{l}{\textit{Proprietary Models}} \\
    GPT-4o~\cite{gpt4o} & 2.0 & 0.0 & 1.5 & 1.3 & 0.0 & 0.7 & 1.0 & 0.0 & 0.6 & 2.1 & 0.0 & 1.2 & 1.1 & 0.0 & 0.9 & 0.0 & 0.0 & 0.0 & 1.3 & 0.0 & \cellcolor{table_color!20}0.8 \\
    Claude Computer Use~\cite{anthropic2024b} & 14.5 & 3.7 & 11.9 & 22.0 & 3.9 & 12.6 & 25.9 & 3.4 & 16.8 & 33.9 & 15.8 & 25.8 & 30.1 & 16.3 & 26.9 & 11.0 & 4.5 & 8.1 & 23.4 & 7.1 & \cellcolor{table_color!20}17.1 \\
    \midrule
    \rowcolor{gray!15}
    \multicolumn{22}{l}{\textit{General Open-source Models}} \\
    Qwen2-VL-7B~\cite{wang2024qwen2} & 0.5 & 0.0 & 0.4 & 2.6 & 0.0 & 1.3 & 1.5 & 0.0 & 0.9 & 6.3 & 0.0 & 3.5 & 3.4 & 1.9 & 3.0 & 0.9 & 0.0 & 0.5 & 2.5 & 0.2 & \cellcolor{table_color!20}1.6 \\
    Qwen2.5-VL-3B~\cite{bai2025qwen2} & - & - & - & - & - & - & - & - & - & - & - & - & - & - & - & - & - & - & - & - & \cellcolor{table_color!20}23.9 \\
    Qwen2.5-VL-7B~\cite{bai2025qwen2} & - & - & - & - & - & - & - & - & - & - & - & - & - & - & - & - & - & - & - & - & \cellcolor{table_color!20}29.0 \\
    Kimi-VL~\cite{team2025kimi} & - & - & - & - & - & - & - & - & - & - & - & - & - & - & - & - & - & - & - & - & \cellcolor{table_color!20}{34.5} \\
    \midrule
    \rowcolor{gray!15}
    \multicolumn{22}{l}{\textit{GUI-specific Models (SFT)}} \\
    SeeClick~\cite{cheng2024seeclick} & 2.5 & 0.0 & 1.9 & 0.6 & 0.0 & 0.3 & 1.0 & 0.0 & 0.6 & 3.5 & 0.0 & 2.0 & 1.1 & 0.0 & 0.9 & 2.8 & 0.0 & 1.5 & 1.8 & 0.0 & \cellcolor{table_color!20}1.1 \\
    CogAgent-18B~\cite{hong2024cogagent} & 7.1 & 3.1 & 6.1 & 14.9 & 0.7 & 8.0 & 9.6 & 0.0 & 5.6 & 22.2 & 1.8 & 13.4 & 13.0 & 0.0 & 10.0 & 5.6 & 0.0 & 3.1 & 12.0 & 0.8 & \cellcolor{table_color!20}7.7 \\
    Aria-UI~\cite{yang2024aria} & 7.6 & 1.6 & 6.1 & 16.2 & 0.0 & 8.4 & 23.7 & 2.1 & 14.7 & 27.1 & 6.4 & 18.1 & 20.3 & 1.9 & 16.1 & 4.7 & 0.0 & 2.6 & 17.1 & 2.0 & \cellcolor{table_color!20}11.3 \\
    OS-Atlas-4B~\cite{atlas} & 2.0 & 0.0 & 1.5 & 7.1 & 0.0 & 3.7 & 3.0 & 1.4 & 2.3 & 9.0 & 5.5 & 7.5 & 5.1 & 3.8 & 4.8 & 5.6 & 0.0 & 3.1 & 5.0 & 1.7 & \cellcolor{table_color!20}3.7 \\
    OS-Atlas-7B~\cite{atlas} & 12.2 & 4.7 & 10.3 & 33.1 & 1.4 & 17.7 & 28.8 & 2.8 & 17.9 & 37.5 & 7.3 & 24.4 & 33.9 & 5.7 & 27.4 & 27.1 & 4.5 & 16.8 & 28.1 & 4.0 & \cellcolor{table_color!20}18.9 \\
    ShowUI-2B~\cite{lin2024showui} & 2.5 & 0.0 & 1.9 & 16.9 & 1.4 & 9.4 & 9.1 & 0.0 & 5.3 & 13.2 & 7.3 & 10.6 & 15.3 & 7.5 & 13.5 & 10.3 & 2.2 & 6.6 & 10.8 & 2.6 & \cellcolor{table_color!20}7.7 \\
    UGround-7B~\cite{gou2024navigating}  & 14.2 & 1.6 & 11.1 & 26.6 & 2.1 & 14.7 & 27.3 & 2.8 & 17.0 & 31.9 & 2.7 & 19.3 & 31.6 & 11.3 & 27.0 & 17.8 & 0.0 & 9.7 & 25.0 & 2.8 & \cellcolor{table_color!20}16.5 \\
    UGround-V1-7B~\cite{gou2024navigating} & - & - & 13.5 & - & - & \underline{35.5} & - & - & 27.8 & - & - & 38.8 & - & - & 48.8 & - & - & \underline{26.1} & - & - & \cellcolor{table_color!20}31.1 \\
    UI-TARS-2B~\cite{qin2025ui} & 17.8 & 4.7 & 14.6 & 47.4 & 4.1 & 26.4 & 42.9 & 6.3 & 27.6 & 56.9 & 17.3 & 39.8 & 50.3 & 17.0 & 42.6 & 21.5 & 5.6 & 14.3 & 39.6 & 8.4 & \cellcolor{table_color!20}27.7 \\
    UI-TARS-7B~\cite{qin2025ui} & 20.8 & \underline{9.4} & {18.0} & \textbf{58.4} & \textbf{12.4} & \textbf{36.1} & \textbf{50.0} & \underline{9.1} & \textbf{32.8} & \textbf{63.9} & \textbf{31.8} & \textbf{50.0} & {63.3} & 20.8 & {53.5} & 30.8 & \textbf{16.9} & 24.5 & {47.8} & \underline{16.2} & \cellcolor{table_color!20}\underline{35.7} \\
    \midrule
    \rowcolor{gray!15}
    \multicolumn{22}{l}{\textit{GUI-specific Models (RL)}} \\
     InfiGUI-R1-3B~\cite{liu2025infigui} & \underline{33.0} & \textbf{14.1} & \underline{28.4} & \underline{51.3} & \textbf{12.4} & 32.4 & \underline{44.9} & 7.0 & \underline{29.0} & 58.3 & {20.0} & {41.7} & \underline{65.5} & \underline{28.3} & \underline{57.0} & \textbf{43.9} & \underline{12.4} & \textbf{29.6} & \underline{49.1} & \underline{14.1} & \cellcolor{table_color!20}\underline{35.7} \\
    UI-R1-3B~\cite{lu2025ui} & 11.2 & 6.3 & - & 22.7 & 4.1 & - & 27.3 & 3.5 & - & 42.4 & 11.8 & - & 32.2 & 11.3 & - & 13.1 & 4.5  & - & - & - & \cellcolor{table_color!20}17.8 \\
    GUI-R1-3B~\cite{xia2025gui} & {26.4} & 7.8 & - & 33.8 & {4.8} & - & 40.9 & 5.6 & - & \underline{61.8} & 17.3 & - & 53.6 & 17.0 & - & 28.1 & 5.6 & - & - & - & \cellcolor{table_color!20}- \\
    GUI-R1-7B~\cite{xia2025gui} & 23.9 & 6.3 & - & 49.4 & {4.8} & - & 38.9 & {8.4} & - & 55.6 & 11.8 & - & 58.7 & {26.4} & - & \underline{42.1} & \textbf{16.9} & - & - & - & \cellcolor{table_color!20}- \\
    \midrule
    \rowcolor{gray!15}
    \multicolumn{22}{l}{\textit{Ours}} \\
     \textbf{GUI-G1-3B} & \textbf{39.6} & \underline{9.4} & \textbf{32.2} & {50.7} & \underline{10.3} & 31.1 & 36.6 & \textbf{11.9} & 26.6 & \underline{61.8} & \underline{30.0} & \underline{48.0} & \textbf{67.2} & \textbf{32.1} & \textbf{59.1} & 23.5 & 10.6 & 16.1 & \textbf{49.5} & \textbf{16.8} & \cellcolor{table_color!20}\textbf{37.1} \\
    \bottomrule
    \end{tabular*}}
\end{table*}

% CAD: 39.59 9.38 32.18
% Dev: 50.65 10.34 31.10
% Creative: 36.63 11.86 26.55
% Scientific: 61.81 30.00 48.03
% Office: 67.23 32.08 59.13
% OS: 23.53 10.64 16.05
% 49.54 16.76 37.10

\section{Related Work}\label{sec:related_work}

\paragraph{Grounding for GUI Agents.} Grounding is central to GUI Agents research~\cite{wang2024gui,zhou2025chop}, driving advancements in data collection and model architecture. Early works, like VUT~\cite{li2021vut} and Spotlight~\cite{lispotlight}, focused on aligning task structures and modalities (e.g., screenshots, instructions) using BERT-based~\cite{devlin2019bert} representations. RUIG~\cite{zhang2023reinforced} used reinforcement learning to map instructions to UI coordinates.With the rise of MLLMs, the focus shifted to fine-tuning pretrained models across platforms for better interaction and adapting to GUI visuals. ShowUI~\cite{lin2024showui} optimized GUI image processing by reducing redundant tokens, improving efficiency. Ferret-UI 2~\cite{li2024ferret} enhanced GUI image understanding through high-resolution encoding and cross-platform adaptability. In contrast, Aria-UI~\cite{yang2024aria} introduces a multi-turn grounding model with sequential reasoning, enabling dynamic multi-step interaction beyond single-shot grounding. More recently, OS-Atlas~\cite{atlas} and UGround~\cite{gou2024navigating} have advanced the field by creating large, open-source datasets and training models that can handle out-of-distribution tasks, achieving state-of-the-art results in GUI grounding. Unlike these approaches, which rely on large datasets and supervised fine-tuning, our work explores how minimal data and an R1-Zero-Like training method can unlock MLLM grounding capabilities for GUI tasks.

\paragraph{R1-Zero-Like Training for MLLMs.} DeepSeek-R1-Zero~\cite{guo2025deepseek} introduces a GRPO-based post-training framework that improves reasoning by encouraging structured outputs. This approach has been extended to multimodal models, with Vision-R1~\cite{huang2025vision}, MM-EUREKA~\cite{meng2025mm}, and VisualThinker-R1-Zero~\cite{zhou2025r1} demonstrating improved performance in vision-language and multimodal reasoning tasks. LMM-R1~\cite{peng2025lmm} applied a two-stage RL method, achieving strong results with low computational costs. However, recent work~\cite{li2025think} challenged and showed that for multimodal classification tasks, reasoning-averse models can outperform reasoning-based ones, suggesting that reasoning is not universally beneficial across tasks. In GUI agents, studies like UI-R1~\cite{lu2025ui}, GUI-R1~\cite{xia2025gui}, and InfiGUI-R1~\cite{liu2025infigui} demonstrated the effectiveness of R1-Zero-Like training in action prediction and grounding. These approaches have shown significant improvements in performance on GUI grounding benchmarks (ScreenSpot~\cite{cheng2024seeclick} and ScreenSpot-Pro~\cite{li2025screenspot}) and the AndroidControl benchmark~\cite{li2024effects}. In this work, we focus solely on the grounding task in GUI agents and explore whether the original settings for R1-Zero-like training in the grounding task are reasonable for GUI scenarios.

\section{Conclusion}\label{sec:conclusion}
% In this work, we revisit the current R1-Zero-Like training setup for R1-style GUI agents on grounding tasks, from the perspectives of templates, reward functions, and the GRPO algorithm.  We first show that longer text thinking can degrade grounding performance, and propose a no-thinking template for training. We then find and analyze opposing types of reward hacking issues in existing reward designs and introduce a box-size constraint to mitigate them, leading to improved performance. Finally, we examine GRPO’s length and difficulty biases in grounding tasks and address them by removing length normalization and incorporating difficulty-based weighting. With these improvements, we train \textbf{GUI-G1-3B} on just 10K samples under the revised GRPO setting. The resulting model surpasses prior state-of-the-art models—including larger ones and the best-performing R1-based approaches—on both the ScreenSpot and ScreenSpot-Pro benchmarks.

In this work, we revisit the current R1-Zero-Like training setup for R1-style GUI agents on grounding tasks, from the perspectives of input design, output evaluation, and the policy update. We first show that longer text thinking can degrade grounding performance, and propose a Fast Thinking Template for training. We then find and analyze opposing types of reward hacking issues in existing reward designs and introduce a box-size constraint to mitigate them, leading to improved performance. Finally, we examine GRPO’s length and difficulty biases in grounding tasks and address them by removing length normalization and incorporating difficulty-based weighting. With these upgrades, our \textbf{GUI-G1-3B}, trained on just 17K samples, outperforms larger and other R1-style models on both ScreenSpot and ScreenSpot-Pro, all while requiring fewer tokens and training stages.
\bibliographystyle{abbrv}
\bibliography{main}

\newpage

\appendix
\section{Acknowledgements}
We thank Yuchong Sun for helpful discussions and insights during the development of this work.

\section{Limitations}\label{sec:limitations}
While our method demonstrates strong performance in GUI grounding, there remain several limitations that offer directions for future work: \textbf{($1$) Focus on grounding.} The current work focuses on grounding, which is essential for GUI agents, but does not cover tasks like action prediction or long-horizon planning. Future research can extend this approach to support full decision-making in GUI interaction. \textbf{($2$) Scope of RL analysis.} The study mainly examines online reinforcement learning, especially GRPO. Other factors such as dataset composition, model design, and hyperparameter tuning are not fully explored and deserve further analysis. \textbf{($3$) Limited training data.} Our model is trained on a relatively small set of public datasets, which constrains its performance ceiling. In future work, we plan to scale up training using larger and more diverse datasets, such as those adopted in GUI-R1~\cite{xia2025gui}, to further improve generalization and robustness.

\section{Analysis Experiments Settings}\label{sec:anlysis_settings}

\subsection{Training Details}
We fine-tune all parameters of the Qwen2.5-VL-3B-Instruct model using samples evenly drawn from three domains: mobile (UIBERT~\cite{bai2021uibert}), web, and desktop (OS-Atlas~\cite{atlas}). Due to computational constraints, we randomly sample 300 grounding examples from each domain. Despite the relatively small dataset, the model achieves strong performance after fine-tuning, ensuring the reliability of our subsequent analysis. Training follows the default setup of the VLM-R1 repository~\cite{shen2025vlm}, using 8 rollouts per example and no KL-divergence regularization by default.

\subsection{Evaluation Metrics}\label{sec:evaluation_metrics}
Specifically, we use the \textbf{relative box size} to measure the size of the predicted bounding box. It is calculated as:
$$
\lambda = \frac{\hat{y}_2 + \hat{x}_2 - \hat{y}_1 - \hat{x}_1}{\texttt{IMAGE\_WIDTH} + \texttt{IMAGE\_HEIGHT}},
$$ 

where $\texttt{IMAGE\_WIDTH}$ and $\texttt{IMAGE\_HEIGHT}$ denote the pixel width and height of the input image, respectively.

\subsection{Difficulty-Aware Weighting Strategy}\label{sec:difficulty_weight}

To compute the difficulty weight $w_q$ for each sample based on its relative box size $\lambda_q \in (0, 1]$, we first take the inverse of the size to reflect the intuition that smaller boxes are harder: $\lambda'_q = \frac{1}{\lambda_i}$. We then normalize the inverted values to the range $[0, 1]$ by computing $\tilde{\lambda}_q = \frac{\lambda'_q - \min_i \lambda'_i}{\max_i \lambda'_i - \min_i \lambda'_i}$. Finally, we linearly rescale the normalized scores to the interval $(0.5, 1.5]$ to obtain the final difficulty weights: $w_q = 0.5 + \tilde{\lambda}_q$. Putting everything together, the final formula is
$$
w_q = 0.5 + \frac{\frac{1}{\lambda_q} - \min_i \left( \frac{1}{\lambda_i} \right)}{\max_i \left( \frac{1}{\lambda_i} \right) - \min_i \left( \frac{1}{\lambda_i} \right)}.
$$
This ensures that harder samples (with smaller boxes) receive higher weights, while keeping the values in a stable and bounded range.

\section{Experiments Details}\label{sec:exp_details}
\subsection{Training Data Composition}\label{sec:training_data}
To provide a comprehensive grounding resource across diverse platforms, we construct a dataset containing 17K samples distributed across three representative domains: Mobile, Web, and Desktop.

\begin{itemize}[leftmargin=20pt,topsep=-3pt,itemsep=3pt]
    \item The Mobile domain is derived from the UI-BERT dataset~\cite{bai2021uibert}, which consists of user interface data collected from Android applications.
    \item The Web domain is sourced from OS-Atlas~\cite{atlas}, including interactive web elements and browser-based environments.
    \item The Desktop domain is also based on OS-Atlas, but focuses on native applications and interfaces from major desktop operating systems, including Windows, Linux, and MacOS.
\end{itemize}

Each domain contains a balanced set of grounding instances that pair natural language commands with corresponding UI elements. Table~\ref{tab:dataset_statistics} summarizes the number of samples collected in each domain.

\begin{table}[h]
\centering
\small
\caption{Statistics and sources of the grounding dataset across five platforms.}\label{tab:dataset_statistics}
\resizebox{\textwidth}{!}{
\begin{tabular}{c|c|c|c|c|c}
\toprule
& \textbf{Mobile} & \textbf{Web} & \textbf{Windows} & \textbf{Linux} & \textbf{MacOS} \\
\midrule\midrule

\textbf{Source} & UI-BERT~\cite{bai2021uibert} & OS-Atlas~\cite{atlas} & OS-Atlas~\cite{atlas} & OS-Atlas~\cite{atlas}
& OS-Atlas~\cite{atlas} \\
\midrule 
\textbf{Size} & 575 & 7,832 & 5,576 & 1,667 & 1,835 \\

\bottomrule 
\end{tabular}
}
\end{table}

\subsection{Output Token Efficiency Analysis}\label{sec:output_tokens}
To assess the efficiency of different models during inference, we compare the average number of output tokens generated per example across Mobile, Desktop, and Web domains on ScreenSpot~\cite{cheng2024seeclick}. As shown in Table~\ref{tab:output_tokens}, \textbf{GUI-G1-3B} generates substantially fewer tokens than \textbf{InfiGUI-R1-3B}~\cite{liu2025infigui} in all domains—approximately one-third as many on average—while maintaining or even improving task accuracy. This compact output not only reduces computational cost but also reflects the model's ability to produce precise and concise responses without relying on verbose intermediate reasoning.

\begin{table}[h]
\centering
\small
\caption{Average number of output tokens generated per example on ScreenSpot during inference.}~\label{tab:output_tokens}
\resizebox{0.5\textwidth}{!}{
\begin{tabular}{l|c|c|c}
\toprule
\textbf{Model} & \textbf{Mobile} & \textbf{Desktop} & \textbf{Web} \\
\midrule\midrule
\textbf{InfiGUI-R1-3B}~\cite{liu2025infigui} & 107 & 107 & 114 \\
\midrule 
\textbf{GUI-G1-3B} & 37 & 39 & 39 \\
\bottomrule 
\end{tabular}
}
\end{table}

\section{Broader Impacts}\label{sec:broader_impacts}

Our work contributes to the development of more robust and accurate GUI agents by addressing key training challenges in reinforcement learning for visual grounding. This could improve the reliability of accessibility tools and human-computer interaction systems. However, care should be taken when deploying such agents in real-world systems, as reward design choices may cause unintended behavior such as reward hacking or bias toward easy cases. We encourage future research to further study fairness, robustness, and privacy considerations in GUI agent training and deployment.

\end{document}